\documentclass[sigconf]{acmart}
\setcopyright{none}
\renewcommand\footnotetextcopyrightpermission[1]{}
\usepackage{subcaption}
\usepackage{multirow}
\AtBeginDocument{}

\copyrightyear{2027}
\acmYear{2027}
\acmDOI{}
\acmISBN{}
\acmConference[KDD '27]{ACM SIGKDD Conference on Knowledge Discovery and Data Mining}
  {August 2027}{San Jose, CA, USA}

\begin{document}

\title{Zenith: Scaling up Ranking Models for Billion-scale Livestreaming Recommendation}

\author{Ruifeng Zhang$^{*\dagger1}$, Zexi Huang$^{*\ddagger2}$, Zikai Wang$^{*3}$, Ke Sun$^{*4}$\\ Bohang Zheng$^{3}$, Yuchen Jiang$^{4}$, Zhe Chen$^{4}$, Zhen Ouyang$^{3}$, Huimin Xie$^{3}$, Phil Shen$^{3}$\\ Junlin Zhang$^{3}$, Yuchao Zheng$^{4}$, Wentao Guo$^{3}$, Qinglei Wang$^{3}$}

\affiliation{
\country{NC State University$^{1}$, TikTok$^{2}$, ByteDance$^{3}$, ByteDance AML$^{4}$\\rzhang38@ncsu.edu, zexi.huang@tiktok.com, \\
\{wangzikai.kevin, ke.sun1, zhengbohang, jiangyuchen.jyc, chenzhe.john, ouyangzhen, xiehuimin.weyman, phil.shen, zhangjunlin.neicul, zhengyuchao.yc, wentao.guo, wangqinglei\}@bytedance.com}}

\thanks{* Equal Contributions. \\ $\dagger$ Work done during internship at TikTok. \\ $\ddagger$ Corresponding author.}

\renewcommand{\shortauthors}{Zhang$^*$, Huang$^*$, Wang$^*$, Ke$^*$ et al.}

\begin{abstract}
Accurately capturing feature interactions is essential in recommender systems, and recent trends show that scaling up model capacity could be a key driver for next-level predictive performance. While prior work has explored various model architectures to capture multi-granularity feature interactions, relatively little attention has been paid to efficient feature handling and scaling model capacity without incurring excessive inference latency.
In this paper, we address this challenge by presenting Zenith, a scalable and efficient ranking architecture that learns complex feature interactions with minimal runtime overhead. Zenith uses Sparse Mixture-of-Experts (SMoE) as a common capacity-scaling mechanism across feature interaction and long-sequence encoding, while adapting its parameterization to each token regime. The Feature Interaction Unit applies tokenwise attention and SMoE to a small, fixed set of high-dimensional Prime Tokens. In contrast, the Sequence Encoding Unit processes thousands of temporal tokens with SMoE-based query, key, and value projections and an SMoE-based feed-forward network. For the livestreaming-specific delayed-feedback problem, Zenith's Target Modeling Unit efficiently connects timely but incomplete short-horizon supervision with complete long-horizon supervision. Its real-world effectiveness is demonstrated through deployment on TikTok Live, a leading online livestreaming platform that attracts billions of users globally. Our online A/B test achieves gains of 9.93\% in Quality Watch Session / User and 8.11\% in Quality Watch Duration / User.

\end{abstract}




\maketitle

\section{Introduction}

Recommender systems play a vital role in modern online platforms, helping users discover relevant content, products, or services. Feature interaction has been recognized as a key mechanism in recommendation models. In recent years, deep learning has emerged as a powerful paradigm for modeling such interactions~\cite{naumov2019dlrm,wang2021dcn,wang2017dcnv1,zhang2023wukong,xu2023dhen}. By learning hierarchical and non-linear representations of users, items, and contextual information, deep neural networks are able to capture complex feature interactions and high-order correlations. Accurate prediction of user--item interaction scores remains the core objective of recommender systems.

With the availability of increasingly large-scale user behavior datasets and the demonstrated effectiveness of scaling laws in large language models~\cite{kaplan2020scaling}, recent research has investigated scaling recommender systems---particularly ranking models---to better predict interaction patterns. Some efforts target the expansion of sparse feature embeddings~\cite{sparsescaling1,sparsescaling2}, while others concentrate on enhancing dense model components or sequence modeling layers above the embedding~\cite{ardalani2022understandingscalinglaw,zhang2023wukong}.

Despite these advancements, deep learning-based recommendation systems still face critical challenges as model size increases. First, many neural architectures, such as transformers, were originally designed for tasks such as natural language processing. Although effective in their native domains, they are not directly optimized for the unique characteristics of the recommendation setting---particularly the increasingly large and diverse sparse feature embeddings that require careful token-level handling. HiFormer~\cite{gui2023hiformer} has shown that the vanilla attention mechanism is not expressive enough for ranking models, but further investigation is needed. Second, most existing methods do not provide flexible designs for scaling up. Instead, they can increase only the number of layers or the number of hidden units~\cite{wang2021dcn,zhang2023wukong,xu2023dhen}. This brute-force scaling method usually leads to diminishing returns and complicates deployment across different feature sets and frameworks. Third, real-world large-scale online recommender systems need to handle millions of requests simultaneously while maintaining millisecond-level latency, making it difficult to retain the benefits of complex architectures without incurring prohibitive computational costs for traditional models.

To address these challenges, we propose \textbf{Zenith} (Optimi\underline{z}ed \underline{E}fficient \underline{N}eural \underline{I}nteraction with \underline{T}okenwise \underline{H}andling), a novel ranking model architecture that enriches model expressiveness without compromising online performance. The key rationale behind Zenith is to treat a few high-dimensional feature tokens as fundamental processing units. Specifically, our initial embedding layer projects the input sparse features into separate embeddings and combines feature embeddings within the same category (e.g., user demographic features and user interaction histories) into higher-dimensional \textbf{Prime Tokens} with a significantly lower token count than the number of raw features. To handle these Prime Tokens, we design two dedicated components: the \textbf{Token Fusion} and \textbf{Token Boost} modules. The Token Fusion module mirrors a cross network by modeling pairwise interactions across the input tokens, while the Token Boost module processes each token independently, preserving and refining its individual semantics. Zenith further coordinates feature interaction with Perceiver-based sequence encoding and a cascaded dataflow for multi-horizon prediction targets. Sparse Mixture-of-Experts (SMoE) is the common scaling primitive in the feature-interaction and sequence-encoding units, but the two units deliberately use different parameterizations. The small, fixed Prime-Token set permits token-specific QKV and expert parameters. A behavior sequence instead contains thousands of variable-length temporal tokens, making per-position parameterization prohibitively large and poorly aligned with position semantics. We therefore share the sequence router and expert banks across temporal positions and introduce Sparse QKV, which applies conditional experts directly to the attention projections in addition to the FFN. For the livestreaming-specific delayed-feedback challenge~\cite{ms3m}, the Target Modeling Unit connects timely but incomplete short-horizon labels with complete long-horizon labels, allowing target-side capacity to scale without backpropagating delayed-label gradients through the shared backbone. Together, these designs adapt sparse capacity scaling to the distinct computational structure of each unit. Their effectiveness is demonstrated through offline and online evaluations on the TikTok Live recommendation platform, which serves billions of users globally.

To summarize, the main contributions of this paper are:
\begin{itemize}
\item We propose \textbf{Zenith}, a novel heterogeneity-aware ranking model architecture. Built upon \textbf{Prime Tokens}, Zenith introduces \textbf{Token Fusion} and \textbf{Token Boost} modules for efficient interactions while preserving token heterogeneity. The architecture coordinates scaling across feature interactions, long user behavior sequences, and multi-horizon prediction targets, using sparse expert activation as a common capacity-scaling mechanism.
\item We introduce Tokenwise Multi-Head Self-Attention (TMHSA) and Tokenwise Sparse Mixture-of-Experts (TSMoE) for the small, fixed Prime-Token set. For long user history sequences, we apply position-shared SMoE to the Perceiver's QKV projections and FFNs. These designs scale model capacity without making the sequence parameters grow with the maximum history length. 
\item Livestreaming produces delayed stay-time feedback at multiple horizons. We introduce a \textbf{Target Modeling Unit} that combines timely short-horizon supervision with complete long-horizon supervision and scales per-horizon capacity without excessive overhead.
\item We present offline evaluations and online A/B test results based on billions of real-world user interactions from TikTok Live recommendation platform. Compared with the production baseline, Zenith improves Quality Watch Session / User by +9.93\% and Quality Watch Duration / User by +8.11\%, demonstrating its practical effectiveness in production.
\end{itemize}

\section{Related Work}

\noindent\textbf{Dense-Model Scaling and Feature Interaction.}
Classical recommender systems combine implicit and explicit feature interactions through architectures such as Deep\&Wide~\cite{deepwide}, DeepFM~\cite{guo2017deepfm}, DLRM~\cite{naumov2019dlrm}, xDeepFM~\cite{lian2018xdeepfm}, and DCN~\cite{wang2017dcnv1,wang2021dcn}. Attention-based models further represent features as tokens and learn their interactions with self-attention~\cite{song2019autoint,li2020transformer1,li2021acn,xu2023dhen,zhang2023wukong,gui2023hiformer}. HiFormer~\cite{gui2023hiformer} introduces heterogeneous attention for recommendation tokens, while Wukong~\cite{zhang2023wukong} and DHEN~\cite{xu2023dhen} scale interaction capacity by increasing width, depth, or module composition. Sparse Mixture-of-Experts (MoE) offers a complementary route: it increases model capacity while activating only a subset of experts~\cite{fedus2022switch,lepikhin2020gshard,du2022glam}. RankMixer focuses on efficient interaction through multi-head token mixing and expert-style transformations~\cite{zhu2025rankmixer}, whereas Zenith uses token-identity-specific projections and experts to preserve heterogeneous Prime-Token representations across deeper layers.

\noindent\textbf{Long-Sequence Encoding.}
Sequential recommenders and joint sequence--interaction models commonly apply attention over retained behavior tokens~\cite{sun2019bert4rec,zeng2025interformer,zhang2026onetrans}. OneTrans is the closest joint architecture, but its retained sequential-query regime is about 10 times shorter than the 10K-scale raw histories considered here. Zenith therefore separates long-sequence encoding from feature interaction: a Perceiver~\cite{jaegle2021perceiver} compresses the history, and SMoE scales its QKV projections and FFN without making the parameter count grow with the sequence length.

\noindent\textbf{Generative Recommenders.}
Another line of work focuses on generative recommenders (GRs)~\cite{googletiger, zhai2024actions, deng2025onerec, xiaohongshugenrank, googleplum, meituanegav2}. Unlike traditional recommender systems that predict user--item interaction scores, GRs directly generate future items based on a user's behavior sequence. Despite recent advances, GRs are unlikely to fully replace traditional approaches. In practice, GRs face significant challenges in training efficiency and inference latency due to the generative modeling process~\cite{hou2025survey}. Moreover, conventional recommender models can complement GRs by serving as reward models to guide generation~\cite{deng2025onerec}. Therefore, GRs represent a parallel research direction and are beyond the scope of this work.

\begin{figure*}[t]
  \centering
  \includegraphics[width=0.95\textwidth]{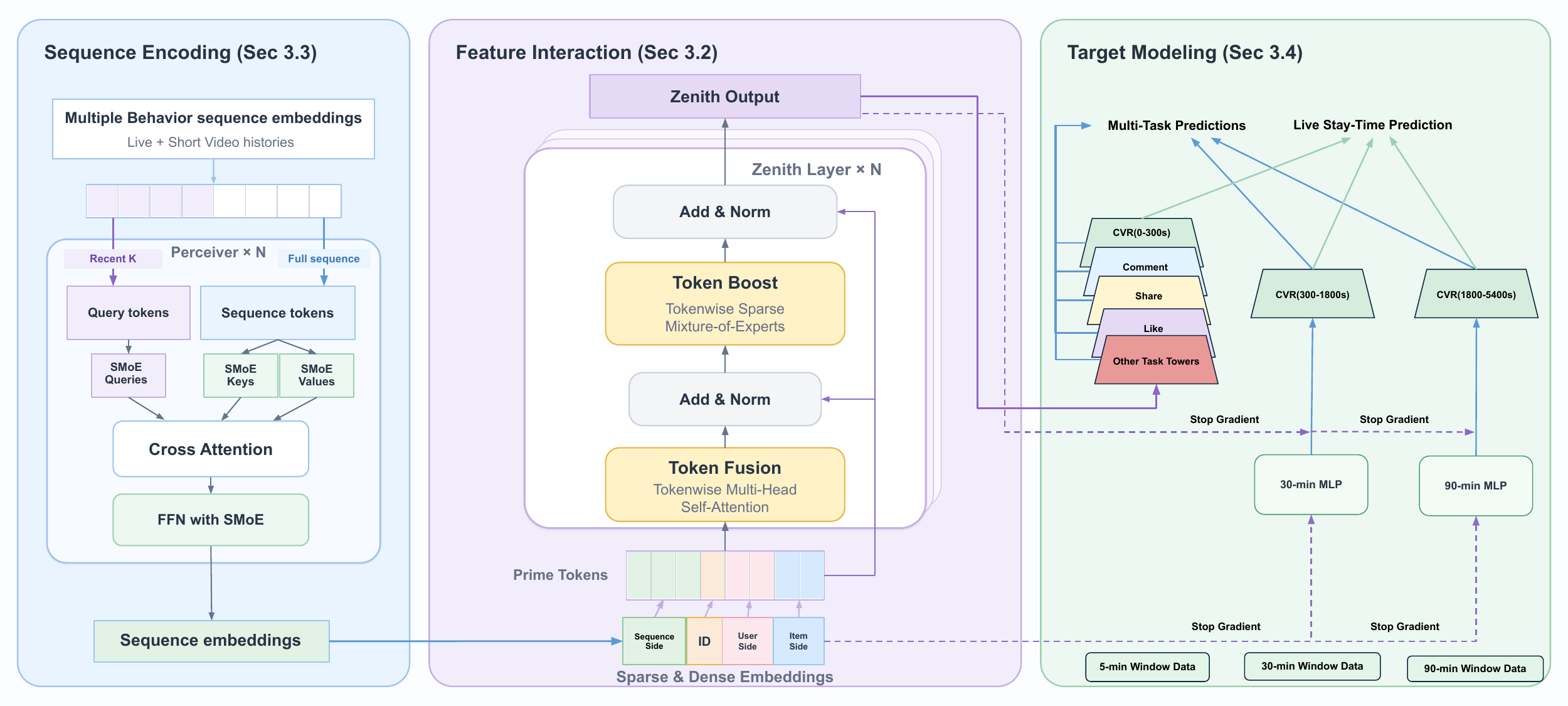}
  \caption{Overview of Zenith's architecture.}
  \Description{Overview of the three-part Zenith architecture.}
  \label{fig:unified_zenithpp}
\end{figure*}

\section{Method}

\subsection{Design Overview}
As shown in Figure~\ref{fig:unified_zenithpp}, Zenith organizes model capacity into three units and uses sparse activation as its central scaling mechanism. The \textbf{Sequence Encoding Unit} (Sec.~\ref{sec:sequence-scaling}) encodes long cross-domain behavior histories with a Perceiver enhanced by SMoE. The \textbf{Feature Interaction Unit} (Sec.~\ref{sec:dense-scaling}) maps feature embeddings into a small, fixed set of Prime Tokens and applies Token Fusion and Token Boost. The \textbf{Target Modeling Unit} (Sec.~\ref{sec:target-scaling}) connects predictions at multiple stay-time horizons through cascaded backbone sharing and gradient isolation. Together, the three units provide scalable sequence, feature-interaction, and target-side capacity.

\subsection{Feature Interaction}
\label{sec:dense-scaling}
\subsubsection{Prime tokenization}
Prior work~\cite{zhang2023wukong,xu2023dhen} has demonstrated the effectiveness of feature tokenization. For our livestreaming recommendation setting, we found that a smaller number of higher-dimensional \textit{Prime Tokens} works better than using one or more tokens for each input feature. Consequently, after standard feature bucketization and embedding lookups, feature embeddings are mapped into a global token space using multi-layer perceptrons (MLPs). Features with similar semantics (e.g., user demographic features) are first concatenated and then mapped to a single Prime Token. Only important features, such as user ID and sequence features, are mapped to separate tokens. As a result, $K$ input features are mapped to $T$ tokens, where $T \ll K$. Unlike traditional methods that rely on explicit feature interactions, the Token Fusion module automatically learns to filter and aggregate relevant token interactions during training. This approach significantly reduces resource consumption compared with traditional models, which often compute explicit interactions for all feature pairs. Moreover, it enables us to use more complex tokenwise methods to improve the representation of each Prime Token. For non-sequence features, our tokenization strategies are as follows:
\begin{enumerate}
    \item \textbf{ID features are assigned separate tokens.} Embeddings of the ID features (e.g., user ID and item ID) are informative and critical for most interaction pairs, so each of them is mapped to a dedicated token.
    \item \textbf{Tokens should preserve complete feature embeddings.} A single feature embedding should not be divided across multiple tokens. Preserving the integrity of feature embeddings is essential for effective token fusion.
    \item \textbf{Remaining tokens should be grouped together based on semantics with balanced information load.} For non-ID features, we group embeddings such that each token contains a similar number of semantically similar features, ensuring balanced information distribution across tokens.
    \item \textbf{Distinct sequence histories are mapped to different tokens.}
    We use several user behavior sequences (e.g., watch and like histories) from both the Live and Video domains, and each sequence is mapped to one Prime Token.
\end{enumerate}

\subsubsection{Zenith Layers}
Zenith scales feature interactions through the \textbf{Token Fusion (TF)} and \textbf{Token Boost (TB)} modules. Because the number and semantics of Prime Tokens are fixed and small, the Feature Interaction Unit can specialize its projections, routers, and experts by token identity. This tokenwise parameter scaling is specific to the feature-interaction regime and is not used for the thousands of temporal positions in the Sequence Encoding Unit. The Token Fusion module functions similarly to conventional feature interaction mechanisms, with the key difference that each Prime Token in our design may aggregate information from multiple features. It performs both explicit inter-token crossing and implicit intra-token feature fusion, enabling the model to selectively focus on the most relevant interactions during training.

The Token Boost module applies tokenwise transformations to improve model expressiveness and enhance token heterogeneity, which, as we demonstrate later in Section~\ref{sec:tok_het}, is key to efficient scaling. The Prime Token design enables the tokens to capture rich and diverse information, collectively functioning as specialized experts akin to those in a Mixture-of-Experts network. Accordingly, the Token Boost module uses tokenwise parameterization, with a separate Token Boost network for each token.

\begin{figure}[t]
  \centering
  \begin{subfigure}[t]{0.49\linewidth}
    \centering
    \includegraphics[width=\linewidth]{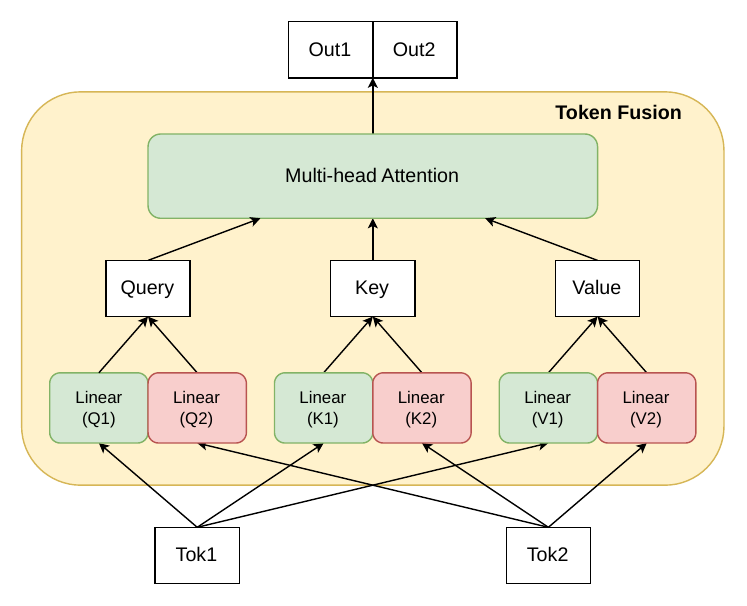}
    \caption{Token Fusion (TMHSA).}
    \label{fig:tf}
  \end{subfigure}\hfill
  \begin{subfigure}[t]{0.49\linewidth}
    \centering
    \includegraphics[width=\linewidth]{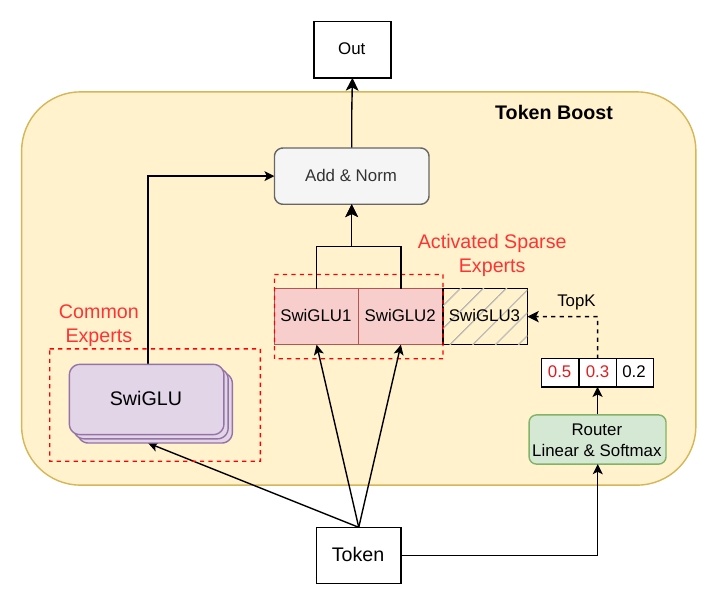}
    \caption{Token Boost (TSMoE).}
    \label{fig:tb}
  \end{subfigure}
  \caption{Designs of the Token Fusion and Token Boost modules for Zenith.}
  \label{fig:zenith_layers}
\end{figure}

\paragraph{Token Fusion: Tokenwise Multi-Head Self-Attention (TMHSA)} \label{sec:tf}
For Zenith, we upgrade the Fused Self-Attention network to Tokenwise Multi-Head Self-Attention for token fusion. TMHSA significantly enhances model expressiveness by assigning distinct query, key, and value projection matrices to each Prime Token, thereby allowing the model to host substantially more parameters. Unlike traditional attention mechanisms that share projection weights across tokens, tokenwise attention introduces unique projections per token, enabling richer interactions without a substantial increase in computational cost. As TMHSA provides sufficient capacity for Prime Tokens to fully interact and exchange information, we do not require a retokenization module in Zenith.

Specifically, denoting the $i$-th token as $t_i$, the output of the multi-head attention layer is computed as:
\vspace{-0.4\baselineskip}
\begin{equation}
\begin{aligned}
O_{TF} &= \mathrm{concat}\!\left(\left\{
\frac{Q_h K_h^\top}{\sqrt{d_k}} V_h
\right\}_{h=1}^H\right), \\
(Q_h, K_h, V_h) &=
\mathrm{concat}\!\left(\{ t_i (q_{(i,h)}, k_{(i,h)}, v_{(i,h)}) \}_{i=1}^n \right).
\end{aligned}
\end{equation}
Here, $H$ is the number of attention heads, and $d_k = d / H$ is the dimensionality of each head. The projection weights $q_{(i,h)}, k_{(i,h)}, v_{(i,h)} \in \mathbb{R}^{d \times d_k}$ are token- and head-specific.
After the residual connection, the input to Token Boost is:
\vspace{-0.4\baselineskip}
\[
X_{TB} = \text{Norm}(O_{TF} + X)
\]

\paragraph{Token Boost: Tokenwise Sparse Mixture-of-Experts (TSMoE)} \label{sec:tb}
To further improve the scaling efficiency of the model, Zenith replaces the standard feedforward neural network with a Tokenwise Sparse Mixture-of-Experts, inspired by the standard Sparse Mixture-of-Experts~\cite{fedus2022switch, du2022glam}. As illustrated in \autoref{fig:tb}, this module comprises two types of experts: \emph{shared} experts, which are always involved in the computation, and \emph{sparse} experts, which are selectively activated based on token-specific routing decisions. This design enables significantly greater model capacity without additional computational overhead.

Let $E_c$ be the number of shared (common) experts, $E_s$ the total number of sparse experts, and $E_a$ the number of sparse experts activated per token. For each token $t_i$, a gating mechanism determines which experts to activate:
\vspace{-0.4\baselineskip}
\[
\text{Gate}(t_i) = t_i W_0^i
\]
where $W_0^i \in \mathbb{R}^{d \times E_s}$ is the routing weight matrix specific to token $t_i$. The top-$E_a$ elements of $\text{Gate}(t_i)$ determine the activated expert indices.

The output of the MoE-enhanced FFN for token $t_i$ is:
\vspace{-0.4\baselineskip}
\[
\text{TB}(t_i) = \sum_{e_c=1}^{E_c} \text{SwiGLU}^{e_c}(t_i) + \sum_{e_a \in \text{TopK}(\text{Gate}(t_i), E_a)} \text{SwiGLU}^{e_a}(t_i)
\]
Here, $\text{SwiGLU}^{e_c}$ and $\text{SwiGLU}^{e_a}$ denote the $e_c$-th shared expert and the $e_a$-th activated sparse expert, respectively, each implemented using the tokenwise SwiGLU function. After each token passes through the Sparse MoE, we denote the Token Boost output by $O_{TB}$.
\looseness=-1

Finally, the module output is passed through a residual connection followed by Layer Normalization:
\vspace{-0.4\baselineskip}
\[
O_{\text{Zenith}} = \text{Norm}(O_{TB} + X)
\]

\subsubsection{Training and Inference Optimizations}
While the dense tokenwise operations improve modeling flexibility, they also inflate the total number of matrix multiplications, reducing throughput on modern accelerators. Sparse MoE in both regimes adds two further complications: (i) expert-activation imbalance, where a minority of experts receive most traffic, and (ii) the overhead of routing computation solely to the selected experts. We tackle these issues with several complementary strategies, detailed in the following sections.

\paragraph{Efficient Tokenwise Computations.}
To eliminate the kernel-launch overhead that accrues when each Prime Token triggers an independent matrix multiplication in the Feature Interaction Unit, we adopt the \emph{GroupedGEMM} primitive available in NVIDIA’s cuBLAS library~\cite{Hejazi2024GroupedGEMM}. GroupedGEMM packs the $N_{\text{tok}}$ tokenwise GEMM operations into a \emph{single} batched call, yielding markedly higher GPU occupancy and end-to-end throughput for the tokenwise design.
\looseness=-1

\paragraph{Learning-Rate Warm-up.}
Early in training, the Sparse-MoE router tends to channel most tokens through only one or two experts, depriving the others of gradient signals. We counteract this collapse with an extended learning-rate warm-up (LR warm-up): training starts at just 0.1\% of the base learning rate and increases linearly to the nominal value over the first one million steps. The resulting low-gradient regime gives the randomly initialized router time to explore diverse routing patterns, after which expert loads become well balanced and every expert receives sufficient training data.
\looseness=-1

\paragraph{Auxiliary Losses.}
To further encourage a balanced workload among experts, we introduce two auxiliary loss terms. The first is the \textit{load-balancing loss}~\cite{fedus2022switch}, which penalizes the mismatch between the router's output probabilities and the actual number of tokens routed to each expert:
\vspace{-0.4\baselineskip}
\[
\mathcal{L}_{\text{load}} = \alpha \cdot \frac{1}{BTE_s} \sum_{i=1}^{E_s} \left( \frac{1}{E_a} f_i \right) \bar{\pi}_i
\]
where
\vspace{-0.4\baselineskip}
\[
f_i = \frac{1}{TB} \sum_{b=1}^{B} \sum_{t=1}^{T} m_{b,t,i}
\]
is the fraction of tokens actually routed to expert $i$, and
\vspace{-0.4\baselineskip}
\[
\bar{\pi}_i = \frac{1}{TB} \sum_{b=1}^{B} \sum_{t=1}^{T} \pi_{b,t,i}
\]
is the average routing probability assigned to expert $i$.

To further prevent expert collapse (i.e., one expert dominating the routing decisions), we employ a second term---referred to as the \textit{z-loss}~\cite{zloss}---to discourage the router from producing overly large logits:
\vspace{-0.4\baselineskip}
\[
\mathcal{L}_{z} = \beta \cdot \frac{1}{BT} \sum_{b=1}^{B} \sum_{t=1}^{T} \left[ \ln \left( \sum_{i=1}^{E_s} e^{z^{b,t,i}} \right) \right]^2
\]
where \( z^{b,t,i} \) denotes the router's output for expert \( i \) at position \( (b, t) \).

\subsection{Sequence Encoding}
\label{sec:sequence-scaling}
The Sequence Encoding Unit converts a long behavior history into a compact sequence embedding before feature interaction. Given the sequence-token embeddings \(X_u=[x_1,\ldots,x_L]\), a Perceiver~\cite{jaegle2021perceiver} uses \(K\) queries, where \(K\ll L\), to read and compress the full sequence. Instead of randomly initialized latent queries, we use the user's \(K\) most recent actions as hard queries:
\vspace{-0.4\baselineskip}
\begin{equation}
Z_u=\operatorname{Perceiver}_1
\left(X_u,\;X_u^{\mathrm{recent}\text{-}K}\right),
\qquad Z_u\in\mathbb{R}^{K\times d},
\label{eq:sequence-compression}
\end{equation}
where \(X_u^{\mathrm{recent}\text{-}K}\) contains the embeddings of the \(K\) most recent actions. A final candidate-specific query then reads the compressed tokens:
\vspace{-0.4\baselineskip}
\begin{equation}
z_{u,c}
=\operatorname{Attention}
\left(q_c,\;K(Z_u),\;V(Z_u)\right),
\qquad z_{u,c}\in\mathbb{R}^{1\times d}.
\label{eq:item-conditioned-sequence}
\end{equation}
A final Perceiver layer constructs \(q_c\) from the embedding of the candidate item. The resulting candidate-conditioned sequence embedding is inserted into the Feature Interaction Unit as a sequence-derived Prime Token.

To scale the Perceiver without densely activating all parameters, we apply SMoE to its attention projections. Unlike the small, fixed Prime-Token set, a behavior sequence contains around \(10\mathrm{K}\) actions. Assigning separate weights to every temporal position would make the parameter count grow with the maximum history length and attach parameters to positions whose semantics vary across users. The Sequence Encoding Unit therefore shares its router and expert banks across temporal positions, whereas the Feature Interaction Unit uses token-specific projections and routers for its small, fixed Prime-Token set. For instance, letting \(\mathcal{A}_{\ell,i}^{Q}\) denote the routed experts selected for query \(r_{\ell,i}\), the sparse query projection is
\vspace{-0.4\baselineskip}
\begin{equation}
\begin{aligned}
\mathcal{A}_{\ell,i}^{Q}
&=\operatorname{TopK}(W^{Q}_{\ell,g}r_{\ell,i},E_a),\\
q_{\ell,i}
&=\alpha^{Q}_{\ell,i,s} f^{Q}_{\ell,s}(r_{\ell,i})
+\sum_{e\in\mathcal{A}_{\ell,i}^{Q}}
\alpha^{Q}_{\ell,i,e} f^{Q}_{\ell,e}(r_{\ell,i}),
\end{aligned}
\label{eq:sequence-sparse-q}
\end{equation}
where \(f^{Q}_{\ell,s}\) is the shared expert, \(f^{Q}_{\ell,e}\) is routed expert \(e\), and \(W^{Q}_{\ell,g}\) is a position-shared router. The mixture weights \(\alpha^Q\) include the adaptive expert ratio defined below. The key and value projections use the same sparse construction. The FFN follows the same position-shared SMoE pattern with SwiGLU experts, analogous to Token Boost but without per-token parameters.

\paragraph{Expert Renorm}
The relative contribution of the shared and routed experts directly affects router learning. A fixed mixture can underuse the routed experts, whereas unconstrained router scores can change both the mixture scale and its shared-to-routed ratio. To achieve stable expert learning, we introduce Expert Renorm with a learnable positive coefficient \(c_{\ell,m}\), where \(m\) identifies an attention or FFN module. Given nonnegative scores \(g_{\ell,i,e}^{(m)}\) for the active routed experts, we compute
\vspace{-0.4\baselineskip}
\begin{equation}
\begin{aligned}
G_{\ell,i}^{(m)}
&=\sum_{e\in\mathcal{A}_{\ell,i}^{(m)}}g_{\ell,i,e}^{(m)},\\
\alpha_{\ell,i,s}^{(m)}
&=\frac{1}{G_{\ell,i}^{(m)}+1},
\quad
\alpha_{\ell,i,e}^{(m)}
=\frac{c_{\ell,m}g_{\ell,i,e}^{(m)}}{G_{\ell,i}^{(m)}+1},\\
h_{\ell,i}^{(m)}
&=\alpha_{\ell,i,s}^{(m)}f_{\ell,s}^{(m)}(x_{\ell,i})
+\sum_{e\in\mathcal{A}_{\ell,i}^{(m)}}
\alpha_{\ell,i,e}^{(m)}f_{\ell,e}^{(m)}(x_{\ell,i}).
\end{aligned}
\label{eq:sequence-expert-renorm}
\end{equation}
The denominator establishes a stable shared-versus-routed reference scale, while \(c_{\ell,m}\) adaptively adjusts the routed contribution. We use this mixture for every sparse \(Q/K/V\) projection in Eq.~\eqref{eq:sequence-sparse-q} and for the SMoE FFN, with module-specific coefficients. During training, the attention coefficient remains close to one, whereas the FFN coefficient grows steadily, indicating that the FFN requires a larger routed-expert contribution. This adaptive ratio improves router learning, as shown by the ablation in Table~\ref{tab:ablation} (more details in Appendix~\ref{app:sequence-design}).
\looseness=-1

\paragraph{KV-Cache-Based Candidate Amortization}\label{sec:kvcache}
Candidates in one ranking request share the same user history but differ in their candidate embeddings. Recomputing the complete Perceiver for every candidate would therefore duplicate most of the sequence computation. During training, we use an asymmetric attention mask: candidate information is excluded from history-state updates, while the candidate query can still attend to the history. Consequently, \(Z_u\), \(K(Z_u)\), and \(V(Z_u)\) depend only on the user and can be reused across candidates.

Serving is factorized into two stages. First, the user-side subgraph runs once to compute \(Z_u\) and cache its projected keys and values. Second, candidate embeddings produce a batch of queries \(q_c\) that cross-attend to this shared cache as in Eq.~\eqref{eq:item-conditioned-sequence}. For \(N_c\) candidates and \(H\) attention heads, the query batch has shape \(N_c\times H\times 1\times d_h\), while the shared key and value caches have shape \(1\times H\times K\times d_h\); a broadcasted batched matrix multiplication reuses the cache without materializing \(N_c\) copies. In our online serving settings with \(N_c = 128\) candidates per request, KV-cache amortization reduces the per-candidate sequence encoder FLOPs to \(1/68\)–\(1/92\) of the unamortized cost, depending on the sequence length.

\subsection{Target Modeling}
\label{sec:target-scaling}

\paragraph{Delayed stay-time supervision}
Beyond fast-feedback targets such as click-through and interactions,
which are observed within seconds and benefit directly from the
feature-interaction and sequence-encoding scaling above, stay time is central to live
recommendation: it is observed on every watch session and directly
reflects user interest. However, it is fully observed only when the
user leaves the live room. To meet both the timeliness and the
accuracy requirements of live recommendation, the production pipeline
emits three delayed streams at 5, 30, and 90 minutes after room
entry, each reporting the stay time observed so far~\cite{ms3m}. For a
user who eventually watches for 40 minutes, the three streams report
5, 30, and 40 minutes: the first two labels are truncated at their
horizons, and only the last is complete. No single stream suffices: a
head trained on the short streams systematically underestimates long
sessions, while one trained only on the 90-minute stream sees
complete labels too late to remain timely. We therefore adopt
the multi-stream segmented formulation
of~\cite{ms3m} and predict per-threshold conversion probabilities on
a grid of $N{=}750$ thresholds $t_1 < \dots < t_N$,
\vspace{-0.4\baselineskip}
\begin{equation}
  \operatorname{CVR}(x) = \Pr(T \ge x), \qquad
  \mathbb{E}[T] \;\approx\; \sum_{i=1}^{N} \bar{w}_i\, \operatorname{CVR}(t_i) + b,
  \label{eq:seg}
\end{equation}
with bin-width weights $\bar{w}_i$. Truncation then acts per
threshold: a sample from horizon $s$ carries unbiased labels exactly
for $x \le s$. Each horizon
$w \in \{\text{5m}, \text{30m}, \text{90m}\}$ owns a tower $g_w$ over
the full grid, trained only on its own stream, and at serving each
threshold is read from the shortest horizon that observes it without
truncation.

\paragraph{Gradient-isolated backbone sharing}
The horizons share the backbone $\Phi$, but the 5-minute stream
provides dense and timely labels while 90-minute positives are sparse
and delayed. We therefore restrict the dataflow to a single
direction: the backbone and the embedding tables are trained by the
fast targets and the 5-minute head, and the long-horizon towers
access shared state only through the stop-gradient operator
$\operatorname{sg}(\cdot)$, so that
\vspace{-0.4\baselineskip}
\begin{equation}
  \frac{\partial \mathcal{L}_w}{\partial \theta_\Phi}
  = \frac{\partial \mathcal{L}_w}{\partial \theta_{\mathrm{emb}}}
  = 0, \qquad w \in \{\text{30m}, \text{90m}\}.
  \label{eq:sg}
\end{equation}
The long-horizon towers reuse the representation learned under dense
supervision, while their gradients cannot perturb it; the
optimization of the backbone is invariant to the tower design. The
towers can thus be modified or enlarged independently and form a
separate axis of scaling.

\paragraph{Scaling the long-horizon towers}
We consider two designs for the long-horizon towers. Both apply a
per-horizon MLP to the concatenation of the frozen backbone output
and a representation extracted from the frozen input tokens, and they
differ in the extraction module $\psi_w$:
\vspace{-0.4\baselineskip}
\begin{equation}
  g_w = \mathrm{MLP}_w\bigl([\operatorname{sg}(h);\,
        \psi_w(\operatorname{sg}(X))]\bigr),
  \quad
  \psi_w =
  \begin{cases}
    \mathrm{MLP}'_w, & \text{lightweight,}\\[2pt]
    \mathrm{Zenith}_w, & \text{replicated,}
  \end{cases}
  \label{eq:towers}
\end{equation}
where $h$ denotes the backbone output, $X$ the input token
embeddings, $\mathrm{MLP}'_w$ a second per-horizon MLP over the token
embeddings, and $\mathrm{Zenith}_w$ an independently parameterized
replica of the backbone. With $\mathrm{MLP}'_w$, extraction is
limited to MLP transformations and adds little online computation.
With $\mathrm{Zenith}_w$, the tower performs full token-level feature
interaction, and each replicated horizon adds roughly the online cost
of the backbone.

\section{Evaluation}

To demonstrate the practical effectiveness of our method, all models are deployed on the TikTok Live recommendation platform.

\subsection{Evaluation Setup}
\subsubsection{Dataset}
We use a production dataset from the ranking stage of the TikTok Live recommendation platform for training and evaluation. The dataset contains real-world user--item interactions collected from a production stream. Its statistics are shown in \autoref{tab:dataset}.
\begin{table}[!t]
\caption{Statistics of the dataset.} \label{tab:dataset}
\centering
\footnotesize
\begin{tabular}{cccc}
\toprule
 & \textbf{\# Instances} & \textbf{\# Features}  & \textbf{\# Targets}\\
\midrule
TikTok Live & 168B & 4552 & 98 \\
\bottomrule
\end{tabular}
\end{table}

\subsubsection{Metrics}
We adopt multiple metrics to comprehensively assess both the computational cost and predictive performance of each model:
\begin{itemize}
    \item \textbf{Model Scale:}
    \begin{itemize}
        \item \textbf{\#Params:} Total number of parameters in the model, including trainable weights from all layers. The sparse embedding table is shared across models in our experiments and excluded from comparison.
        \item \textbf{GFLOPs:} Total number of floating-point operations (in billions) required for a single inference. This provides a more accurate measure of computational load, especially when not all parameters are actively used.
    \end{itemize}
    \item \textbf{Predictive Metrics:}
    \begin{itemize}
    \item \textbf{LogLoss:} The logarithmic loss of the predicted probabilities against the ground truth.
    \item \textbf{AUC:} Area Under the ROC Curve, measuring the model's ability to rank positive samples above negatives.
    \item \textbf{UAUC:} User-Averaged AUC, which averages AUC scores per user to reflect personalized ranking quality.
    \item \textbf{RegAUC:} A pairwise concordance metric for continuous regression targets, measuring how often the prediction preserves the ordering of two samples with different target values.
\end{itemize}
\end{itemize}

All results are reported as both raw values and relative improvements over our baseline model in production.

\subsubsection{Baselines \& Hyperparameters}
Because the three units address different scaling axes, we compare each against the most relevant controlled baselines: DLRMs for feature interaction, Decoder and Perceiver architectures for sequence encoding, and cascaded backbone--tower variants for target modeling.
We choose widely used DLRMs and recent strong baselines, including DCN-V2~\cite{wang2021dcn}, Transformer~\cite{vaswani2017attention}, DHEN~\cite{xu2023dhen}, HiFormer~\cite{gui2023hiformer}, Wukong~\cite{zhang2023wukong}, and RankMixer~\cite{zhu2025rankmixer}. We tune their hyperparameters to obtain the required model scales as follows:
\begin{itemize}
    \item \textbf{DCN-V2}: We scale up the model by increasing the number of layers and using an MLP in the Cross Network instead of a single projection matrix. We apply sequential modeling with a Deep Network after the Cross Network.
    \item \textbf{Transformer}: We scale up the vanilla Transformer by increasing the number of layers and the hidden size.
    \item \textbf{DHEN}: We use a combination of DCN-V2's Cross Network and a multi-head self-attention module. We scale up the model by increasing the number of DHEN layers and the size of the MLPs attached after each interaction module.
    \item \textbf{HiFormer}: We similarly scale up HiFormer by increasing its depth and hidden size.
    \item \textbf{Wukong}: We scale up the model by adding layers and widening the MLP in its FM Block.
    \item \textbf{RankMixer}: We scale up the model by adding layers, widening its FFNs, and increasing the number of experts.
\end{itemize}

Zenith employs 32 Prime Tokens grouped from the original feature embeddings: 3 ID tokens, 13 sequence-derived tokens, and 16 tokens for other features. The dimension of each Prime Token ranges from 256 to 1024, and the number of layers ranges from 2 to 8, depending on the model scale.

To evaluate Zenith's Sequence Encoding Unit, we compare a Transformer decoder~\cite{vaswani2017attention}, a dense Perceiver~\cite{jaegle2021perceiver}, and Perceiver-SMoE under the same Feature Interaction Unit. All sequence encoders have two layers and use a hidden size of 128 and an FFN width of 256.
\subsubsection{Training}
For all models, we use the RMSPropV2 optimizer with a learning rate of 0.01, momentum of 0.99999, and an initialization factor of 0.015625, along with learning rate warm-up (more details in Appendix~\ref{app:training-efficiency}).

\subsection{Results} \label{sec:results}
To the best of our knowledge, no existing baseline provides a directly matched comparison across Zenith's three scaling axes under the same industrial serving constraints. Joint sequence--interaction and generative recommenders operate in different sequence-length or prediction regimes~\cite{zhang2026onetrans,deng2025onerec}; we therefore use unit-wise controlled comparisons.
\subsubsection{Feature Interaction}

\begin{figure}[t]                                                                                                                                           \centering                                                                                                                                          \includegraphics[width=\linewidth]{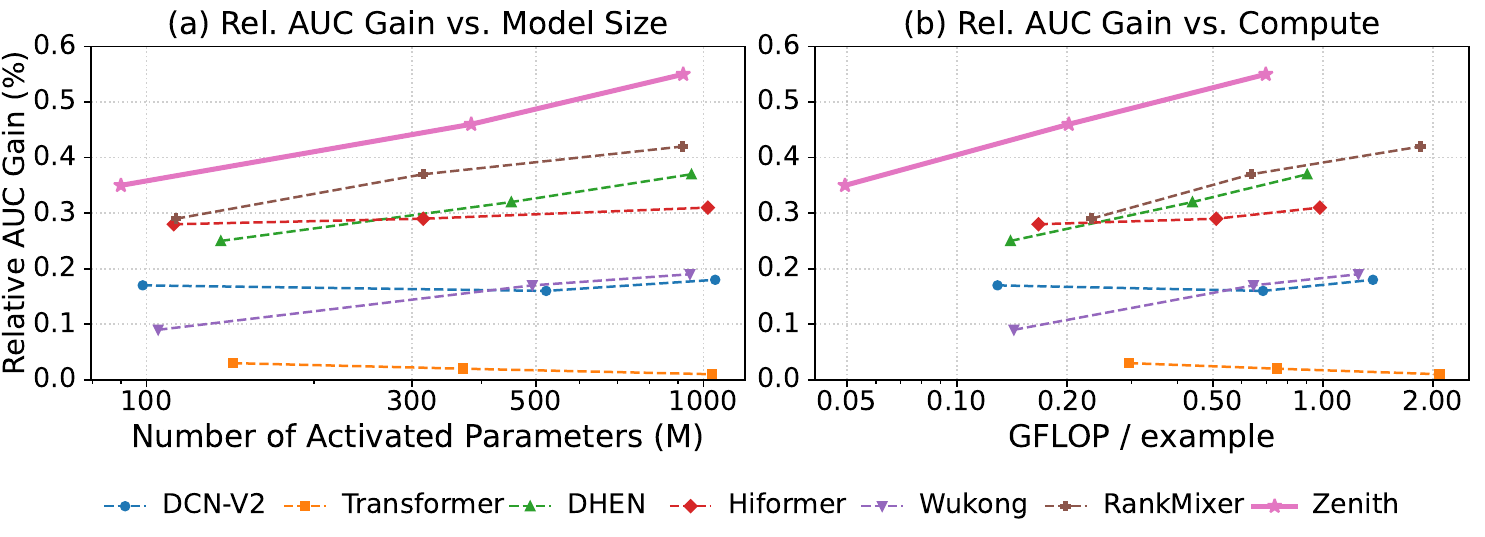}                                                                                \caption{Model performance comparison between our proposed architectures and the baselines. All AUC gains are shown relative to the baseline DCN-V2 model.}                                                                                \label{fig:auc_scaling_combined}
\end{figure}
\begin{table*}[!t]
\caption{Model performance at different scales. Values in parentheses indicate relative improvement over the baseline. Parameter and FLOP counts include only the Feature Interaction Unit.}
\label{tab:offline}
\centering
\footnotesize
\begin{tabular}{c l c c c c c}
\toprule
\textbf{Scale} & \textbf{Architecture} & \textbf{\#Activated Params (M)} & \textbf{GFLOPs (per example)}& \textbf{AUC} & \textbf{Logloss} & \textbf{UAUC} \\
\midrule
Baseline & DCN-V2 & 22.8& 0.0302& .81151 & .08255 & .69627 \\
\midrule
\multirow{7}{*}{Small}
& DCN-V2 & 98.5 & 0.129& .81290 (+0.17\%) & .08237 (-0.22\%) & .69871 (+0.35\%) \\
& Transformer & 143 & 0.295& .81173 (+0.03\%) & .08252 (-0.04\%) & .69652 (+0.04\%) \\
& DHEN & 136 & 0.140& .81351 (+0.25\%) & .08227 (-0.34\%) & .69877 (+0.36\%) \\
& HiFormer & 112 & 0.167& .81378 (+0.28\%) & .08227 (-0.34\%) & .69841 (+0.31\%) \\
& Wukong & 105 & 0.143& .81221 (+0.09\%) & .08247 (-0.10\%) & .69688 (+0.09\%) \\
& RankMixer & 113 & 0.233& .81390 (+0.29\%) & .08223 (-0.39\%) & .69970 (+0.49\%) \\
& \textbf{Zenith} & \textbf{90} & \textbf{0.0494}& \textbf{.81438 (+0.35\%)} & \textbf{.08214 (-0.50\%)} & \textbf{.70036 (+0.59\%)} \\
\midrule
\multirow{7}{*}{Medium}
& DCN-V2 & 522 & 0.686& .81283 (+0.16\%) & .08237 (-0.22\%) & .69881 (+0.36\%) \\
& Transformer & 370 & 0.750& .81168 (+0.02\%) & .08252 (-0.04\%) & .69627 (+0.00\%) \\
& DHEN & 452 & 0.440& .81409 (+0.32\%) & .08219 (-0.44\%) & .69990 (+0.52\%) \\
& HiFormer & 314 & 0.511& .81389 (+0.29\%) & .08227 (-0.34\%) & .69889 (+0.38\%) \\
& Wukong & 493 & 0.646& .81287 (+0.17\%) & .08238 (-0.21\%) & .69822 (+0.28\%) \\
& RankMixer & 314 & 0.637& .81452 (+0.37\%) & .08212 (-0.52\%) & .70045 (+0.60\%) \\
& \textbf{Zenith} & \textbf{383} & \textbf{0.202}& \textbf{.81521 (+0.46\%)} & \textbf{.08202 (-0.64\%)} & \textbf{.70170 (+0.78\%)} \\
\midrule
\multirow{7}{*}{Large}
& DCN-V2 & 1050 & 1.37& .81299 (+0.18\%) & .08235 (-0.24\%) & .69876 (+0.36\%) \\
& Transformer & 1037 & 2.08& .81163 (+0.01\%) & .08251 (-0.05\%) & .69601 (-0.04\%) \\
& DHEN & 952 & 0.906& .81448 (+0.37\%) & .08213 (-0.51\%) & .70036 (+0.59\%) \\
& HiFormer & 1019 & 0.981& .81402 (+0.31\%) & .08226 (-0.35\%) & .69944 (+0.46\%) \\
& Wukong & 946 & 1.25& .81306 (+0.19\%) & .08234 (-0.25\%) & .69850 (+0.32\%) \\
& RankMixer & 918 & 1.85& .81488 (+0.42\%) & .08207 (-0.58\%) & .70110 (+0.69\%) \\
& \textbf{Zenith} & \textbf{920} & \textbf{0.698}& \textbf{.81594 (+0.55\%)} & \textbf{.08188 (-0.81\%)} & \textbf{.70303 (+0.97\%)} \\
\bottomrule
\end{tabular}
\normalsize
\end{table*}

\autoref{fig:auc_scaling_combined} presents an analysis of the relative AUC gain across various model sizes and compute budgets. All baseline models are equipped with the best Sequence Encoding Unit and Target Modeling Unit for fair comparisons. We report results from two perspectives: (a) performance versus model size, and (b) performance versus inference-time FLOPs. Note that for model size, we report the number of activated parameters per inference. The baseline is a 22.8M-parameter DCN-V2 model. Vanilla Transformers do not perform well because they use shared $Q$, $K$, and $V$ projections for all tokens, which limits their ability to capture heterogeneous feature interactions. Although DCN exhibits moderate gains when scaled to 100M parameters, further increasing the number of DCN layers or expanding its MLP does not yield additional improvements. A similar trend is observed for HiFormer, which significantly outperforms DCNs but still fails to scale effectively. In contrast, Wukong scales more effectively than DCN but only outperforms it at scales of 500M parameters or more, and the performance gain remains modest. Wukong employs a large MLP to map interaction results back to tokens, which leads to inefficient parameter utilization—limiting its advantage to very large scales.

DHEN integrates DCN-V2's Cross Network with self-attention, delivering consistently stronger performance than both DCN and Transformer across all scales, with AUC improvements of 0.25\%--0.37\%. DHEN also shows good scalability. However, its benefits remain marginal, and its hybrid architecture increases tuning complexity because of the heterogeneous components involved.

RankMixer provides the strongest baseline in our experiments. Relative to RankMixer, Zenith improves AUC by 0.06\%, 0.08\%, and 0.13\%, reduces Logloss by 0.11\%, 0.12\%, and 0.23\%, and improves UAUC by 0.09\%, 0.18\%, and 0.28\% at the small, medium, and large scales, respectively. Meanwhile, Zenith uses 78.8\%, 68.3\%, and 62.3\% fewer inference GFLOPs per example. These consistent gains at substantially lower compute suggest that Zenith's Feature Interaction Unit scales more efficiently than RankMixer.

\begin{figure}[t]
  \centering
  \begin{subfigure}[t]{0.44\linewidth}
    \centering
    \includegraphics[width=\linewidth]{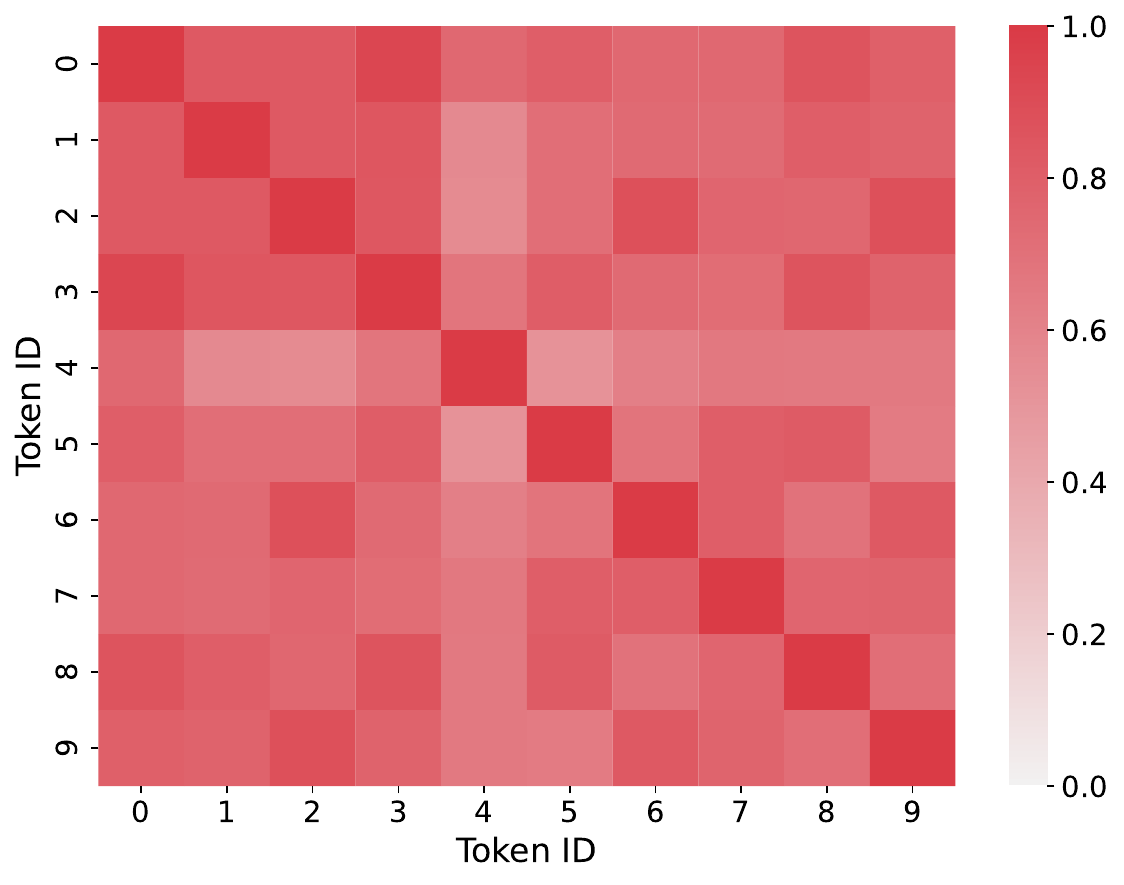}
    \caption{Zenith w/o TSMoE}
    \label{fig:zen+wotsmoe}
  \end{subfigure}\hfill
  \begin{subfigure}[t]{0.44\linewidth}
    \centering
    \includegraphics[width=\linewidth]{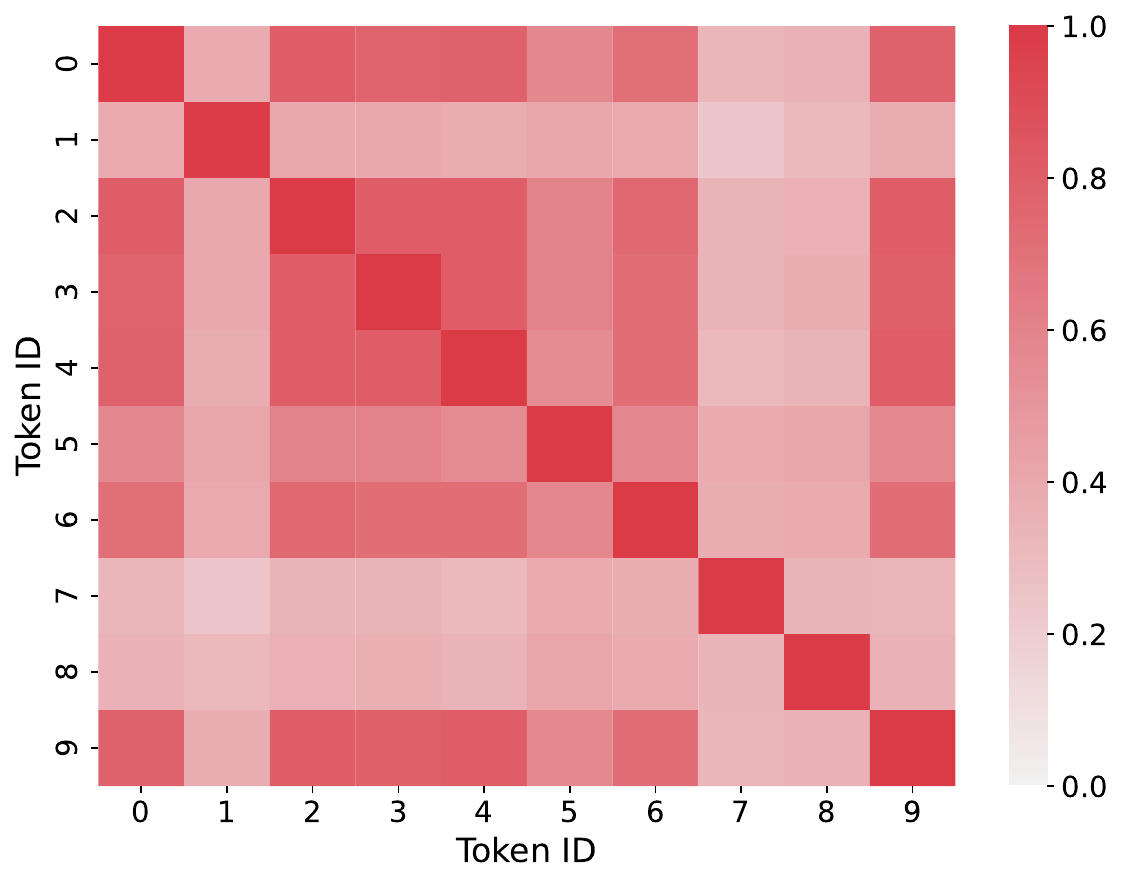}
    \caption{Zenith with TSMoE}
    \label{fig:zen+wtsmoe}
  \end{subfigure}
  \caption{Cosine similarity between output tokens from Zenith with and without TSMoE. }\label{fig:token_sim}
\end{figure}

\subsubsection{Sequence Encoding}
\label{sec:sequence-eval}
\begin{table*}[!t]
\caption{Sequence-encoding performance across architectures and history lengths. All variants use the same Zenith (medium) Feature Interaction Unit. Activated parameters and GFLOPs cover only sequence encoding; GFLOPs are reported per user and amortized across candidate items at serving, details are in Sec.~\ref{sec:kvcache}.}
\label{tab:sequence-scaling}
\centering
\footnotesize
\setlength{\tabcolsep}{2.5pt}
\begin{tabular}{c l c c c c c}
\toprule
\textbf{Max History} & \textbf{Architecture} & \textbf{\#Activated Params (M)} & \textbf{GFLOPs (per user)} & \textbf{AUC} & \textbf{Logloss} & \textbf{UAUC} \\
\midrule
\multirow{4}{*}{1K}
  & Decoder                         & 0.58 & 0.081 & .81297 & .08234 & .69752 \\
  & Perceiver                       & 0.94 & 0.152 & .81346 (+0.06\%) & .08229 (-0.06\%) & .69808 (+0.08\%) \\
  & Perceiver-SMoE ($Q=128$)        & 1.11 & 0.285 & .81394 (+0.12\%) & .08222 (-0.15\%) & .69919 (+0.24\%) \\
  & Perceiver-SMoE ($Q=256$)        & 1.11 & 0.390 & .81419 (+0.15\%) & .08218 (-0.19\%) & .69968 (+0.31\%) \\
\midrule
\multirow{4}{*}{5K}
  & Decoder                         & 0.58 & 0.393 & .81330 (+0.04\%) & .08230 (-0.05\%) & .69801 (+0.07\%) \\
  & Perceiver                       & 0.94 & 0.592 & .81386 (+0.11\%) & .08223 (-0.13\%) & .69892 (+0.20\%) \\
  & Perceiver-SMoE ($Q=128$)        & 1.11 & 1.109 & .81443 (+0.18\%) & .08215 (-0.23\%) & .70017 (+0.38\%) \\
  & Perceiver-SMoE ($Q=256$)        & 1.11 & 1.344 & .81467 (+0.21\%) & .08210 (-0.29\%) & .70073 (+0.46\%) \\
\midrule
\multirow{4}{*}{10K}
  & Decoder                         & 0.58 & 0.784 & .81354 (+0.07\%) & .08227 (-0.09\%) & .69843 (+0.13\%) \\
  & Perceiver                       & 0.94 & 1.144 & .81427 (+0.16\%) & .08217 (-0.21\%) & .69975 (+0.32\%) \\
  & Perceiver-SMoE ($Q=128$)        & 1.11 & 2.146 & .81484 (+0.23\%) & .08208 (-0.32\%) & .70100 (+0.50\%) \\
  & \textbf{Perceiver-SMoE ($Q=256$)}
                                    & \textbf{1.11} & \textbf{2.545} & \textbf{.81521 (+0.28\%)} & \textbf{.08202 (-0.39\%)} & \textbf{.70170 (+0.60\%)} \\
\bottomrule
\end{tabular}
\normalsize
\end{table*}
Table~\ref{tab:sequence-scaling} compares three sequence encoders. The Decoder baseline uses a single candidate-conditioned query to attend directly to the retained history, without a latent compression stage. Both Perceiver and Perceiver-SMoE comprise two layers, with \(Q\in\{128,256\}\) denoting the number of first-layer compression queries. Each SMoE module contains one shared expert that is always active and eight routed experts, of which the top two are selected per instance. The feature-interaction backbone, training data, and feature set are held fixed, and all relative metric changes use the first row (1K Decoder) as the common reference.

At a 1K history length, replacing the Decoder with a dense Perceiver increases the activated parameter count from 0.58M to 0.94M (\(1.62\times\)) and improves AUC by 0.06\%. With only 18\% more activated parameters than the dense Perceiver, Perceiver-SMoE doubles this gain to 0.12\% at \(Q=128\); doubling the query count to \(Q=256\) further increases the gain to 0.15\% without increasing the activated parameter count beyond 1.11M. Across history lengths, the Decoder benefits little from additional context, with relative AUC gains of 0, 0.04\%, and 0.07\% at 1K, 5K, and 10K, respectively. In contrast, Perceiver-SMoE scales monotonically with both history length and query count: the \(Q=128\) variant improves from 0.12\% to 0.18\% and 0.23\%, while the \(Q=256\) variant improves from 0.15\% to 0.21\% and 0.28\%. Logloss and UAUC exhibit the same pattern, showing that Perceiver-SMoE has the best scalability (more details in Appendix~\ref{app:user-cohorts}).

Sequence-encoding GFLOPs are reported per user, whereas Feature Interaction GFLOPs are per user--item pair. The history-side representation and cached keys/values are shared across candidate items, so the sequence cost is amortized over the candidate batch at serving. Details are in Sec.~\ref{sec:kvcache}

\subsubsection{Target Modeling}
\label{sec:target-eval}
\begin{table*}[!t]
  \caption{Target-path scaling. Model cost and RegAUC on all samples
    and on high-stay-time segments; values in parentheses are relative
    improvements over the baseline.}
  \label{tab:target-ladder}
  \centering
  \footnotesize
  \begin{tabular}{@{}lcccccc@{}}
    \toprule
    & \#Activated & GFLOPs & \multicolumn{4}{c}{RegAUC} \\
    \cmidrule(lr){4-7}
    Cascade & Params (M) & (per example) & All & $T{\ge}120$s
      & $T{\ge}300$s & $T{\ge}600$s \\
    \midrule
    DCN-V2{\,+\,}MLP{\,+\,}MLP    & 31.6 & 0.0420
      & 0.6680 & 0.6289 & 0.6123 & 0.6061 \\
    \textbf{Zenith{\,+\,}MLP{\,+\,}MLP}
      & \textbf{98.8} & \textbf{0.0544}
      & \textbf{0.6685 (+0.07\%)} & \textbf{0.6370 (+1.29\%)}
      & \textbf{0.6183 (+0.98\%)} & \textbf{0.6117 (+0.92\%)} \\
    Zenith$\,\times 3$          & 271.0 & 0.1482
      & 0.6689 (+0.13\%) & 0.6382 (+1.48\%)
      & 0.6191 (+1.11\%) & 0.6124 (+1.04\%) \\
    \bottomrule
  \end{tabular}
\end{table*}

We compare three cascaded architectures: (1) the previous production
baseline, DCN-V2{\,+\,}MLP{\,+\,}MLP; (2)
Zenith{\,+\,}MLP{\,+\,}MLP, which replaces the DCN-V2 backbone with
Zenith while retaining the lightweight long-horizon branches; and
(3) Zenith$\,\times 3$, which additionally introduces a separate
Zenith replica in each of the two long-horizon branches
(Details are in Sec.~\ref{sec:target-scaling}). Because stay time is heavy-tailed and
long sessions are sparse but important, we evaluate RegAUC on all
samples and on the subsets with $T \ge 120$ s, $T \ge 300$ s, and
$T \ge 600$ s, using labels from the 90-minute stream.

Both Zenith-based cascades outperform the production baseline overall
and on every long-session subset (Table~\ref{tab:target-ladder}). The
improvements are substantially larger for long sessions: replacing the
DCN-V2 backbone with Zenith improves RegAUC by 1.29\% for
$T \ge 120$ s, compared with 0.07\% over all samples. Because gradients
from the long-horizon losses are stopped at the shared backbone, this
gain reflects the stronger shared representation rather than updates
from the long-horizon branches. Replacing the lightweight branch
extractors with Zenith replicas increases the improvement at
$T \ge 120$ s from 1.29\% to 1.48\%, an additional 0.19 percentage
points, while increasing both activated parameters and per-example
GFLOPs by approximately $2.7\times$. Among the two Zenith-based
variants, Zenith{\,+\,}MLP{\,+\,}MLP therefore provides the better
accuracy--efficiency trade-off, capturing most of the observed gain
at substantially lower computational cost.

\begin{table*}[!t]
\caption{Online performance gains and breakdown (CUPED results with $p<0.01$).}
\label{tab:online_results}
\centering
\footnotesize
\setlength{\tabcolsep}{7pt}
\renewcommand{\arraystretch}{1.05}
\begin{tabular}{lcccc}
\toprule
\textbf{Online Metric}
& \shortstack{\textbf{Zenith}\\\textbf{vs.\ Production Baseline}}
& \shortstack{\textbf{Feature}\\\textbf{Interaction Unit}}
& \shortstack{\textbf{Sequence}\\\textbf{Encoding Unit}}
& \shortstack{\textbf{Target}\\\textbf{Modeling Unit}} \\
\midrule
Quality Watch Session / User  & $+9.93\%$ & $+4.88\%$ & $+2.80\%$ & $+1.42\%$ \\
Quality Watch Duration / User & $+8.11\%$ & $+3.95\%$ & $+1.90\%$ & $+1.50\%$ \\
\bottomrule
\end{tabular}
\end{table*}

\begin{table}[!t]
\centering
\footnotesize
\caption{Relative online serving cost of Zenith.}
\label{tab:serving_cost}
\begin{tabular}{lc}
\toprule
\textbf{Metric} & \textbf{vs. Baseline} \\
\midrule
Model size & 79.91$\times$ \\
Serving GPUs & 2.01$\times$ (after full optimization) \\
Latency & +4.39\% \\
% GPU peak utilization & 3.28$\times$ \\
% SM Tensor Core activity & 3.33$\times$ \\
\bottomrule
\end{tabular}
\end{table}

\subsubsection{Design Analysis} \label{sec:tok_het}
We now provide a quantitative analysis of the specific design choices in Zenith.

First, we highlight the key advantage of tokenwise parameterization in the Token Boost module, which enables Zenith to scale more effectively. Initially, we applied a conventional non-tokenwise SMoE in Token Boost, but encountered performance saturation beyond the 50M-parameter scale. We increased the Prime Token dimension from 512 to 768, the number of Prime Tokens from 8 to 16, and the number of interaction layers from 4 to 6. Although these changes approximately doubled the parameter count, none produced further performance gains. This observation motivated us to examine how tokens interact inside the model. As shown in \autoref{fig:zen+wotsmoe}, without tokenwise parameterization, the Prime Tokens become nearly homogeneous after several interaction layers. Subsequent layers therefore receive increasingly similar token representations and have less distinctive information to process, which is consistent with the observed performance saturation. To address this issue, Prime Tokens should remain heterogeneous after each interaction layer and preserve distinctive information for subsequent feature interaction. TSMoE equips the conventional SMoE with tokenwise parameterization, providing separate capacity for different Prime Tokens to capture different information. As shown in \autoref{fig:zen+wtsmoe}, TSMoE reduces the average absolute cosine similarity between third-layer Prime Tokens from \textbf{0.68} to \textbf{0.47}. The resulting representations remain more distinct across deeper interaction layers, allowing Zenith to continue benefiting from increased model capacity instead of reaching the same early plateau.

We also provide an ablation study to evaluate the impact of key design choices in Zenith.
\begingroup
As summarized in Table~\ref{tab:ablation}, both TSMoE and TMHSA contribute substantially to model performance, while Expert Renorm, learning-rate warm-up, and the auxiliary loss provide additional gains. This confirms that each component plays a positive role, with the architectural designs yielding the most significant improvements. Meanwhile, substituting Prime Tokenization with an MLP-based auto-grouping approach incurs only a 0.09\% AUC degradation, demonstrating that Zenith is robust even without expert-tuned feature tokenization.
\looseness=-1
\emergencystretch=0.5em
\par\endgroup

\begin{table}[!t]
\footnotesize
\caption{Ablation results relative to the corresponding full models. Expert Renorm uses the sequence-encoding setting.}
\label{tab:ablation}
\centering
\begin{tabular}{c c c}
\toprule
\textbf{Ablation Class} & \textbf{Description} & \textbf{AUC ($\Delta$\%)} \\
\midrule
Tokenization
  & Prime Tokenization $\rightarrow$ Auto-grouping & $-0.09$ \\
\midrule
\multirow{2}{*}{Token Fusion}
  & Disable Token Fusion                          & $-0.60$ \\
  & TMHSA $\rightarrow$ Self-Attn                 & $-0.12$ \\
\midrule
\multirow{2}{*}{Token Boost}
  & Disable Token Boost                           & $-0.32$ \\
  & TSMoE $\rightarrow$ MoE                       & $-0.10$ \\
\midrule
\multirow{3}{*}{Optimizations}
  & Remove LR Warmup                              & $-0.04$ \\
  & Remove Auxiliary Loss                         & $-0.01$ \\
  & Remove Expert Renorm                 & $-0.03$ \\
\bottomrule
\end{tabular}
\normalsize
\end{table}
\subsubsection{Online Results and Serving Cost}

To validate the architecture under real-world serving constraints, we conducted a month-long online A/B test comparing the production baseline with Zenith (medium) on TikTok Live. The production baseline combines DCN-V2 for feature interaction with the baseline Sequence Encoding and Target Modeling Units. We report two business metrics for livestreaming recommendation: Quality Watch Session / User and Quality Watch Duration / User. A Quality Watch Session is either a deep watch session (e.g., longer than one minute) or a session accompanied by a positive action such as a like, comment, follow, share, or gift; Quality Watch Duration is the total time spent in such sessions. As shown in \autoref{tab:online_results}, Zenith improves the two metrics by 9.93\% and 8.11\%, respectively. Table~\ref{tab:online_results} further decomposes the gains across the Feature Interaction, Sequence Encoding, and Target Modeling Units. The Feature Interaction Unit contributes the largest individual gain. Jointly scaling all three units yields gains that exceed the sum of their single-unit gains: 9.93\% versus 9.10\% for Quality Watch Session and 8.11\% versus 7.35\% for Quality Watch Duration, suggesting complementary effects among the three units.
Online gains appear larger than offline gains because the latter measure only CTR objective, whereas the deployed ranker jointly optimizes multiple targets. Some sparse auxiliary targets show larger gains, such as over 10\% relative AUC improvement on dislike.

% \autoref{tab:serving_cost} reports the relative online serving cost of deploying Zenith compared to the DCN baseline. With operator-level and system-level optimizations, Zenith scales the model size by 79.91$\times$ while requiring only 7.89$\times$ serving GPUs and increasing inference latency by only 4.39\%. This efficiency is enabled by improved hardware utilization, including 3.28$\times$ higher GPU peak utilization and 3.33$\times$ higher Tensor Core activity.

\autoref{tab:serving_cost} reports the serving cost of Zenith relative to the DCN baseline. With 79.91$\times$ the model size, Zenith requires only 2.01$\times$ as many serving GPUs after serving optimizations, including grouped expert GEMMs with fused bias/activation, dedicated Top-K routing, and dynamic batching. The latency increases by only 4.39\% (less than 1 ms in the serving path), remaining within the serving budget. The greater capacity of a single Zenith model also enables consolidation across objectives, reducing the complete serving stack's cost to about 0.8$\times$ that of the previous multi-model DCN stack (More serving details are in Appendix~\ref{app:production-details}).

\section{Conclusion}

In this work, we present Zenith, a scalable and efficient neural architecture tailored for large-scale recommendation systems. Zenith uses Sparse MoE as a common capacity-scaling mechanism while adapting it to different token regimes. The Feature Interaction Unit uses token-specific parameters for a small, fixed Prime-Token set. The Sequence Encoding Unit instead shares its router and experts across thousands of temporal positions and sparsifies the Q/K/V projections and FFN, allowing sequence capacity to grow without per-position parameterization. Together with the Target Modeling Unit, these designs coordinate feature-interaction, sequence, and target-side scaling under practical serving constraints. We hope that our work will be an important step toward improving user experiences through personalized recommendation.

\bibliographystyle{ACM-Reference-Format}
\bibliography{ref}

@inproceedings{guo2017deepfm,
  title={DeepFM: A factorization-machine based neural network for CTR prediction},
  author={Guo, Huifeng and Tang, Ruiming and Ye, Yunming and Li, Zhenguo and He, Xiuqiang},
  booktitle={Proceedings of the 26th International Joint Conference on Artificial Intelligence (IJCAI)},
  pages={1725--1731},
  year={2017}
}

@article{naumov2019dlrm,
  title={Deep Learning Recommendation Model for Personalization and Recommendation Systems},
  author={Naumov, Maxim and Mudigere, Dheevatsa and Shi, Hao-Jun and Huang, Jianyu and Sundaram, Narayanan and Park, Jongsoo and others},
  journal={arXiv preprint arXiv:1906.00091},
  year={2019}
}

@article{zhang2023wukong,
  title={Wukong: Towards a scaling law for large-scale recommendation},
  author={Zhang, Buyun and Luo, Liang and Chen, Yuxin and Nie, Jade and Liu, Xi and Guo, Daifeng and Zhao, Yanli and Li, Shen and Hao, Yuchen and Yao, Yantao and Lakshminarayanan, Guna and Wen, Ellie Dingqiao and Park, Jongsoo and Naumov, Maxim and Chen, Wenlin},
  journal={arXiv preprint arXiv:2403.02545},
  year={2024}
}

@article{xu2023dhen,
  title={DHEN: A deep and hierarchical ensemble network for large-scale click-through rate prediction},
  author={Zhang, Buyun and Luo, Liang and Liu, Xi and Li, Jay and Chen, Zeliang and Zhang, Weilin and Wei, Xiaohan and Hao, Yuchen and Tsang, Michael and Wang, Wenjun and Liu, Yang and Li, Huayu and Badr, Yasmine and Park, Jongsoo and Yang, Jiyan and Mudigere, Dheevatsa and Wen, Ellie},
  journal={arXiv preprint arXiv:2203.11014},
  year={2022}
}

@inproceedings{deepwide,
  title={Wide \& deep learning for recommender systems},
  author={Cheng, Heng-Tze and Koc, Levent and Harmsen, Jeremiah and Shaked, Tal and Chandra, Tushar and Aradhye, Hrishi and Anderson, Glen and Corrado, Greg and Chai, Wei and Ispir, Mustafa and others},
  booktitle={Proceedings of the 1st workshop on deep learning for recommender systems},
  pages={7--10},
  year={2016}
}

@inproceedings{wang2021dcn,
  title={Dcn v2: Improved deep \& cross network and practical lessons for web-scale learning to rank systems},
  author={Wang, Ruoxi and Shivanna, Rakesh and Cheng, Derek and Jain, Sagar and Lin, Dong and Hong, Lichan and Chi, Ed},
  booktitle={Proceedings of the web conference 2021},
  pages={1785--1797},
  year={2021}
}

@incollection{wang2017dcnv1,
  title={Deep \& cross network for ad click predictions},
  author={Wang, Ruoxi and Fu, Bin and Fu, Gang and Wang, Mingliang},
  booktitle={Proceedings of the ADKDD'17},
  pages={1--7},
  year={2017}
}

@inproceedings{lian2018xdeepfm,
  title={xdeepfm: Combining explicit and implicit feature interactions for recommender systems},
  author={Lian, Jianxun and Zhou, Xiaohuan and Zhang, Fuzheng and Chen, Zhongxia and Xie, Xing and Sun, Guangzhong},
  booktitle={Proceedings of the 24th ACM SIGKDD international conference on knowledge discovery \& data mining},
  pages={1754--1763},
  year={2018}
}

@inproceedings{song2019autoint,
  title={Autoint: Automatic feature interaction learning via self-attentive neural networks},
  author={Song, Weiping and Shi, Chence and Xiao, Zhiping and Duan, Zhijian and Xu, Yewen and Zhang, Ming and Tang, Jian},
  booktitle={Proceedings of the 28th ACM international conference on information and knowledge management},
  pages={1161--1170},
  year={2019}
}

@inproceedings{li2020transformer1,
  title={Interpretable click-through rate prediction through hierarchical attention},
  author={Li, Zeyu and Cheng, Wei and Chen, Yang and Chen, Haifeng and Wang, Wei},
  booktitle={Proceedings of the 13th international conference on web search and data mining},
  pages={313--321},
  year={2020}
}

@article{li2021acn,
  title={Attentive capsule network for click-through rate and conversion rate prediction in online advertising},
  author={Li, Dongfang and Hu, Baotian and Chen, Qingcai and Wang, Xiao and Qi, Quanchang and Wang, Liubin and Liu, Haishan},
  journal={Knowledge-based systems},
  volume={211},
  pages={106522},
  year={2021},
  publisher={Elsevier}
}

@article{kaplan2020scaling,
  title={Scaling laws for neural language models},
  author={Kaplan, Jared and McCandlish, Sam and Henighan, Tom and Brown, Tom B and Chess, Benjamin and Child, Rewon and Gray, Scott and Radford, Alec and Wu, Jeffrey and Amodei, Dario},
  journal={arXiv preprint arXiv:2001.08361},
  year={2020}
}

@article{fedus2022switch,
  title={Switch transformers: Scaling to trillion parameter models with simple and efficient sparsity},
  author={Fedus, William and Zoph, Barret and Shazeer, Noam},
  journal={Journal of Machine Learning Research},
  volume={23},
  number={120},
  pages={1--39},
  year={2022}
}

@article{lepikhin2020gshard,
  title={Gshard: Scaling giant models with conditional computation and automatic sharding},
  author={Lepikhin, Dmitry and Lee, HyoukJoong and Xu, Yuanzhong and Chen, Dehao and Firat, Orhan and Huang, Yanping and Krikun, Maxim and Shazeer, Noam and Chen, Zhifeng},
  journal={arXiv preprint arXiv:2006.16668},
  year={2020}
}

@inproceedings{du2022glam,
  title={Glam: Efficient scaling of language models with mixture-of-experts},
  author={Du, Nan and Huang, Yanping and Dai, Andrew M and Tong, Simon and Lepikhin, Dmitry and Xu, Yuanzhong and Krikun, Maxim and Zhou, Yanqi and Yu, Adams Wei and Firat, Orhan and others},
  booktitle={International conference on machine learning},
  pages={5547--5569},
  year={2022},
  organization={PMLR}
}

@article{gui2023hiformer,
  title={Hiformer: Heterogeneous feature interactions learning with transformers for recommender systems},
  author={Gui, Huan and Wang, Ruoxi and Yin, Ke and Jin, Long and Kula, Maciej and Xu, Taibai and Hong, Lichan and Chi, Ed H},
  journal={arXiv preprint arXiv:2311.05884},
  year={2023}
}

@article{sparsescaling1,
  title={High-performance, distributed training of large-scale deep learning recommendation models},
  author={Mudigere, Dheevatsa and Hao, Yuchen and Huang, Jianyu and Tulloch, Andrew and Sridharan, Srinivas and Liu, Xing and Ozdal, Mustafa and Nie, Jade and Park, Jongsoo and Luo, Liang and others},
  journal={arXiv preprint arXiv:2104.05158},
  year={2021},
  publisher={Apr}
}

@inproceedings{sparsescaling2,
  title={Persia: An open, hybrid system scaling deep learning-based recommenders up to 100 trillion parameters},
  author={Lian, Xiangru and Yuan, Binhang and Zhu, Xuefeng and Wang, Yulong and He, Yongjun and Wu, Honghuan and Sun, Lei and Lyu, Haodong and Liu, Chengjun and Dong, Xing and others},
  booktitle={Proceedings of the 28th ACM SIGKDD Conference on Knowledge Discovery and Data Mining},
  pages={3288--3298},
  year={2022}
}

@article{ardalani2022understandingscalinglaw,
  title={Understanding scaling laws for recommendation models},
  author={Ardalani, Newsha and Wu, Carole-Jean and Chen, Zeliang and Bhushanam, Bhargav and Aziz, Adnan},
  journal={arXiv preprint arXiv:2208.08489},
  year={2022}
}

@misc{Hejazi2024GroupedGEMM,
  author       = {Babak Hejazi},
  title        = {Introducing Grouped {GEMM} APIs in cuBLAS and More Performance Updates},
  howpublished = {\url{https://developer.nvidia.com/blog/introducing-grouped-gemm-apis-in-cublas-and-more-performance-updates/}},
  note         = {NVIDIA Technical Blog},
  year         = {2024},
  month        = jun
}

@article{zhai2024actions,
  title={Actions speak louder than words: Trillion-parameter sequential transducers for generative recommendations},
  author={Zhai, Jiaqi and Liao, Lucy and Liu, Xing and Wang, Yueming and Li, Rui and Cao, Xuan and Gao, Leon and Gong, Zhaojie and Gu, Fangda and He, Michael and others},
  journal={arXiv preprint arXiv:2402.17152},
  year={2024}
}

@article{deng2025onerec,
  title={Onerec: Unifying retrieve and rank with generative recommender and iterative preference alignment},
  author={Deng, Jiaxin and Wang, Shiyao and Cai, Kuo and Ren, Lejian and Hu, Qigen and Ding, Weifeng and Luo, Qiang and Zhou, Guorui},
  journal={arXiv preprint arXiv:2502.18965},
  year={2025}
}

@article{zloss,
  title={St-moe: Designing stable and transferable sparse expert models},
  author={Zoph, Barret and Bello, Irwan and Kumar, Sameer and Du, Nan and Huang, Yanping and Dean, Jeff and Shazeer, Noam and Fedus, William},
  journal={arXiv preprint arXiv:2202.08906},
  year={2022}
}

@article{googletiger,
  title={Recommender systems with generative retrieval},
  author={Rajput, Shashank and Mehta, Nikhil and Singh, Anima and Hulikal Keshavan, Raghunandan and Vu, Trung and Heldt, Lukasz and Hong, Lichan and Tay, Yi and Tran, Vinh and Samost, Jonah and others},
  journal={Advances in Neural Information Processing Systems},
  volume={36},
  pages={10299--10315},
  year={2023}
}

@article{meituanegav2,
  title={Ega-v2: An end-to-end generative framework for industrial advertising},
  author={Zheng, Zuowu and Wang, Ze and Yang, Fan and Fan, Jiangke and Zhang, Teng and Wang, Yongkang and Wang, Xingxing},
  journal={arXiv preprint arXiv:2505.17549},
  year={2025}
}

@article{xiaohongshugenrank,
  title={Towards Large-scale Generative Ranking},
  author={Huang, Yanhua and Chen, Yuqi and Cao, Xiong and Yang, Rui and Qi, Mingliang and Zhu, Yinghao and Han, Qingchang and Liu, Yaowei and Liu, Zhaoyu and Yao, Xuefeng and others},
  journal={arXiv preprint arXiv:2505.04180},
  year={2025}
}

@article{googleplum,
  title={Plum: Adapting pre-trained language models for industrial-scale generative recommendations},
  author={He, Ruining and Heldt, Lukasz and Hong, Lichan and Keshavan, Raghunandan and Mao, Shifan and Mehta, Nikhil and Su, Zhengyang and Tsai, Alicia and Wang, Yueqi and Wang, Shao-Chuan and others},
  journal={arXiv preprint arXiv:2510.07784},
  year={2025}
}

@article{hou2025survey,
  title={A survey on generative recommendation: Data, model, and tasks},
  author={Hou, Min and Wu, Le and Liao, Yuxin and Yang, Yonghui and Zhang, Zhen and Zheng, Changlong and Wu, Han and Hong, Richang},
  journal={arXiv preprint arXiv:2510.27157},
  year={2025}
}

@article{vaswani2017attention,
  title={Attention is all you need},
  author={Vaswani, Ashish and Shazeer, Noam and Parmar, Niki and Uszkoreit, Jakob and Jones, Llion and Gomez, Aidan N and Kaiser, {\L}ukasz and Polosukhin, Illia},
  journal={Advances in neural information processing systems},
  volume={30},
  year={2017}
}

@inproceedings{zhu2025rankmixer,
  title={Rankmixer: Scaling up ranking models in industrial recommenders},
  author={Zhu, Jie and Fan, Zhifang and Zhu, Xiaoxie and Jiang, Yuchen and Wang, Hangyu and Han, Xintian and Ding, Haoran and Wang, Xinmin and Zhao, Wenlin and Gong, Zhen and others},
  booktitle={Proceedings of the 34th ACM International Conference on Information and Knowledge Management},
  pages={6309--6316},
  year={2025}
}

@inproceedings{zhang2026onetrans,
  title={Onetrans: Unified feature interaction and sequence modeling with one transformer in industrial recommender},
  author={Zhang, Zhaoqi and Pei, Haolei and Guo, Jun and Wang, Tianyu and Feng, Yufei and Sun, Hui and Liu, Shaowei and Sun, Aixin},
  booktitle={Proceedings of the ACM Web Conference 2026},
  pages={8162--8170},
  year={2026}
}

@inproceedings{sun2019bert4rec,
  title={BERT4Rec: Sequential recommendation with bidirectional encoder representations from transformer},
  author={Sun, Fei and Liu, Jun and Wu, Jian and Pei, Changhua and Lin, Xiao and Ou, Wenwu and Jiang, Peng},
  booktitle={Proceedings of the 28th ACM international conference on information and knowledge management},
  pages={1441--1450},
  year={2019}
}

@inproceedings{zeng2025interformer,
  title={InterFormer: Effective Heterogeneous Interaction Learning for Click-Through Rate Prediction},
  author={Zeng, Zhichen and Liu, Xiaolong and Hang, Mengyue and Liu, Xiaoyi and Zhou, Qinghai and Yang, Chaofei and Liu, Yiqun and Ruan, Yichen and Chen, Laming and Chen, Yuxin and others},
  booktitle={Proceedings of the 34th ACM International Conference on Information and Knowledge Management},
  pages={6225--6233},
  year={2025}
}

@inproceedings{jaegle2021perceiver,
  title={Perceiver: General Perception with Iterative Attention},
  author={Jaegle, Andrew and Gimeno, Felix and Brock, Andrew and Vinyals, Oriol and Zisserman, Andrew and Carreira, Jo{\~a}o},
  booktitle={Proceedings of the 38th International Conference on Machine Learning},
  series={Proceedings of Machine Learning Research},
  volume={139},
  pages={4651--4664},
  year={2021},
  publisher={PMLR},
  url={https://proceedings.mlr.press/v139/jaegle21a.html}
}

@inproceedings{ms3m,
  title={Towards Unbiased and Real-Time Staytime Prediction for Live Streaming Recommendation},
  author={Zhao, Haiyuan and Zhang, Changshuo and Wang, Yang and Wang, Hao and Ouyang, Zhen and Yuan, Bin and Wang, Qinglei and Liu, Zuotao},
  booktitle={Proceedings of the 34th ACM International Conference on Information and Knowledge Management},
  pages={6293--6300},
  year={2025},
  publisher={Association for Computing Machinery},
  address={New York, NY, USA},
  doi={10.1145/3746252.3761570},
  url={https://doi.org/10.1145/3746252.3761570}
}

\clearpage \appendix \begingroup \section{Additional Implementation Details and Evaluation}
\label{app:additional}

\subsection{Additional Sequence-Encoder Design Comparisons}
\label{app:sequence-design}

\paragraph{Recent-action versus learned queries.}
With the same 10K-history model, replacing the $K$ most recent actions with learned clustering queries reduces both AUC and UAUC by 0.01\%. Recent-action queries are therefore a simple, slightly better choice in this comparison, rather than the main source of the sequence encoder's gain.

\paragraph{Routing and shared experts.}
Directly applying standard SMoE to long behavior histories introduces two difficulties: adding shared experts halves the observed gain, while using sigmoid routing scores directly with the load-balancing losses causes routing collapse. The attention implementation addresses routing overhead with a fused Q/K/V router shared across layers, avoiding scattered per-layer routing operations. Routing scores are decoupled by purpose: softmax scores are used for load balancing, while sigmoid scores weight expert outputs. Expert Renorm, defined in Eq.~\eqref{eq:sequence-expert-renorm}, learns the relative contribution of shared and routed experts. Together, these sequence-specific designs match the predictive gain obtained by doubling a dense model's width while keeping FLOPs nearly unchanged.

\subsection{History Length and User Activity}
\label{app:user-cohorts}

The controlled history-length comparison in Table~\ref{tab:sequence-scaling} fixes the Zenith-medium Feature Interaction backbone, data, and features. With $Q=256$, the AUC/UAUC improvements over the common 1K Decoder reference increase from $+0.15\%/+0.31\%$ at 1K to $+0.21\%/+0.46\%$ at 5K and $+0.28\%/+0.60\%$ at 10K. These controlled results show the benefit of extending the available history.

Online results for Zenith-medium show that the gains vary across engagement metrics. In the 1K rollout, Valuable Watch Live Days changed by $+0.92\%$, $+1.45\%$, $+1.25\%$, and $+0.50\%$ for users with 0, 1--7, 8--14, and 15+ active days, respectively; the marginal gain is largest before the most-active cohort saturates.

In the 10K rollout, the cohort with 0 active days gained $+0.28\%$ AUC, $+0.48\%$ UAUC, $+1.3\%$ penetration, and $+0.5\%$ duration, while the cohort with 1--7 active days gained $+1.5\%$ penetration and $+0.7\%$ duration. These production rollouts bundle other model changes and are not included in the controlled results in Table~\ref{tab:sequence-scaling}. They show that long history primarily benefits users whose sequences reach the cap, whereas low-activity gains can also come from stronger feature interaction, cross-domain history, and generalization.

\subsection{Training Efficiency}
\label{app:training-efficiency}

The original Zenith-medium training configuration requires 24 standard GPU cards, compared with 8 for the production baseline. Regularizing fragmented computation into attention, FFN, and batched GEMM operations, together with expert parallelism, mixed precision, FlashAttention, operator fusion, sequence merging, padding removal, and user-level reuse, reduces the final training requirement to 8 cards, matching the baseline. Streaming multiprocessor activity (SMA) increases from about 2\% to about 50\%.

The measured throughput improvements are $+30\%$ from expert parallelism, $+81.8\%$ from mixed precision, FlashAttention, and operator fusion, and $+56.5\%$ from sequence and data-path optimizations. These component-level improvements are not additive and should not be summed into an end-to-end speedup.

\subsection{Production Evaluation and Deployment Economics}
\label{app:production-details}

\paragraph{Online protocol and metric interpretation.}
The main online experiment runs continuously for one month, with treatment and control assigned at random and each containing tens of millions of users. The reported gains in Table~\ref{tab:online_results} are CUPED-adjusted and statistically significant at $p<0.01$. The offline $+0.78\%$ CTR UAUC improvement is meaningful for the production baseline, but measures only the primary CTR objective. The deployed model also affects stay time, interactions, and sparse positive and negative feedback before downstream score fusion. Consequently, the offline CTR UAUC change does not numerically predict the online business-metric changes. The unit-wise online comparisons in Table~\ref{tab:online_results} provide the direct evidence for contributions from all three units.

\paragraph{Additional online baseline comparisons.}
At comparable serving costs, Zenith improves watch sessions per user and watch duration per user by approximately 2\% and 1.5\% over DCN-V2-medium, and by 1.8\% and 1.4\% over RankMixer, respectively. OneTrans achieves comparable online performance to Zenith while requiring 50\% more serving resources. These comparisons complement the controlled offline results. With return on investment defined as online metric gain per unit of serving cost, Zenith achieves approximately $25\times$ the ROI of the strategy that combines the production DCN-V2 ranker with a larger candidate pool.

\paragraph{Serving optimization breakdown.}
The original single-model configuration uses $7.89\times$ the serving GPUs of a single DCN baseline. This configuration already includes sparse expert activation and sequence caching, but precedes the later serving optimizations. In this original measurement, GPU peak utilization and Tensor Core activity are $3.28\times$ and $3.33\times$ those of the baseline, respectively.

Grouped expert GEMMs with fused bias and activation improve expert throughput by 40\%. A dedicated Top-K routing operator reduces routing time from about 7\,ms to tens of microseconds. Mixed precision, FlashAttention, and graph fusion reduce graph latency by about 50\%, and dynamic batching improves per-GPU throughput by 54\%. These gains overlap. The measured end-to-end serving-cost reduction is about 75\%, consistent with the optimized $2.01\times$ GPU ratio reported in Table~\ref{tab:serving_cost}.

The single-model and complete-stack cost ratios use different denominators. The $2.01\times$ ratio compares one optimized Zenith model with one DCN model. The previous production stack instead uses separate DCN models for different objectives. Zenith's greater capacity enables those objectives to be served by one unified model, reducing the complete stack's cost to about $0.8\times$ that of the previous multi-model stack. The 4.39\% latency increase is less than 1\,ms in the serving path and remains within the service budget; no user-experience or reliability concern is observed during the month-long A/B test.
 \endgroup \end{document}